\documentclass[letterpaper, 10 pt, conference]{ieeeconf}  

\IEEEoverridecommandlockouts                              

\usepackage{graphics} 
\usepackage{epsfig} 
\usepackage{amsmath} 
\usepackage{amssymb}  

\usepackage{subfigure}
\usepackage[table]{xcolor}
\usepackage{multirow}
\usepackage{multicol}
\usepackage{algorithm} 
\usepackage[noend]{algpseudocode}
\usepackage[acronym]{glossaries}
\usepackage{hyperref}

\title{\LARGE \bf
Preventing Errors in Person Detection: A Part-Based Self-Monitoring Framework}

\author{Franziska Schwaiger$^{1}$, Andrea Matic$^{1}$, Karsten Roscher$^{1}$, and Stephan G\"unnemann$^{2}$ 
\thanks{This work has been supported by the German Federal Ministry for Economic Affairs and Climate Action as part of the safe.trAIn project and by the Bavarian Ministry of Economic Affairs, Regional Development and Energy with funds from the Hightech Agenda Bayern.}
\thanks{$^{1}$ All authors are with the Fraunhofer Institute for Cognitive Systems IKS in Munich, Germany. E-Mail: $<$firstname$>$.$<$lastname$>$@iks.fraunhofer.de}%
\thanks{$^{2}$ Stephan G\"unnemann is with the Technical University Munich (TUM), Germany. E-Mail: stephan.guennemann@tum.de}%
}

\newacronym{tp}{true positive}{true positive}
\newacronym{fp}{false positive}{false positive}
\newacronym{fn}{false negative}{false negative}
\newacronym{gt}{GT}{Ground Truth}
\newacronym{sm}{SMF}{Self-Monitoring Framework}
\newacronym{mcc}{MCC}{Matthew Correlation Coefficient}
\newacronym{iou}{IOU}{Intersection Over Union}
\newacronym{map}{mAP}{Mean Average Precision}
\newacronym{auroc}{AUROC}{Area under the Receiver Operating Characteristic}
\newacronym{aupr}{AUPR}{Area under the Precision Recall Curve}
\newacronym{voc}{PASCAL VOC}{Pascal Visual Object Classes}
\newacronym{coco}{MS-COCO}{Microsoft Common Objects in Context}
\newacronym{fpn}{FPN}{Feature Pyramid Network}
\newacronym{rcnn}{C-RCNN}{Cascade R-CNN}
\newacronym{dpm}{DPM}{Deformable Part Model}
\newacronym{hog}{HOG}{Histogram of Oriented Gradients}

\newcommand\MyBoxvvvv[4]{%
  \fbox{\lower0.75cm
  \cellcolor[HTML]{#4}
    \vbox to 0.8cm{\vfil
      \hbox to 0.8cm{\hfil\parbox{1.4cm}{%
           #1 \par
            #2 \par
             #3
      }\hfil}
      \vfil}%
  }%
}

\newcommand\MyBox[4]{%
    \cellcolor[HTML]{#4}
    \begin{tabular}{c}\\#1\\#2\\\vspace{5pt}#3\\\end{tabular}
}
 
 \newcommand\MyBoxLeft[1]{
  \vbox to 0.8cm{\vfil
    \hbox to 0.8cm{\hfil
      \begin{minipage}{1.4cm}
        \vspace{11pt}
        #1
        \vspace{11pt}
      \end{minipage}
    \hfil}
    \vfil}%
}

\newcommand\MyBoxRight[0]{
  \vbox to 0.8cm{\vfil
    \hbox to 0.8cm{\hfil
      \begin{minipage}{0.6cm}
        B 
        MD 
        SD
      \end{minipage}
    \hfil}
    \vfil}%
}

\newcommand\MyBoxTop[1]{
  \mbox{#1}
 }

\begin{document}

\maketitle
\thispagestyle{empty}
\pagestyle{empty}


\begin{abstract}

The ability to detect learned objects regardless of their appearance is crucial for autonomous systems in real-world applications. Especially for detecting humans, which is often a fundamental task in safety-critical applications, it is vital to prevent errors. To address this challenge, we propose a self-monitoring framework that allows for the perception system to perform plausibility checks at runtime. We show that by incorporating an additional component for detecting human body parts, we are able to significantly reduce the number of missed human detections by factors of up to 9 when compared to a baseline setup, which was trained only on holistic person objects. Additionally, we found that training a model jointly on humans and their body parts leads to a substantial reduction in false positive detections by up to 50\% compared to training on humans alone. We performed comprehensive experiments on the publicly available datasets DensePose and Pascal VOC in order to demonstrate the effectiveness of our framework. Code is available at \textcolor{cyan}{\url{https://github.com/FraunhoferIKS/smf-object-detection}}.

\end{abstract}


\section{INTRODUCTION}

\begin{figure}[tb]
    \centering
    \includegraphics[width=0.45\textwidth]{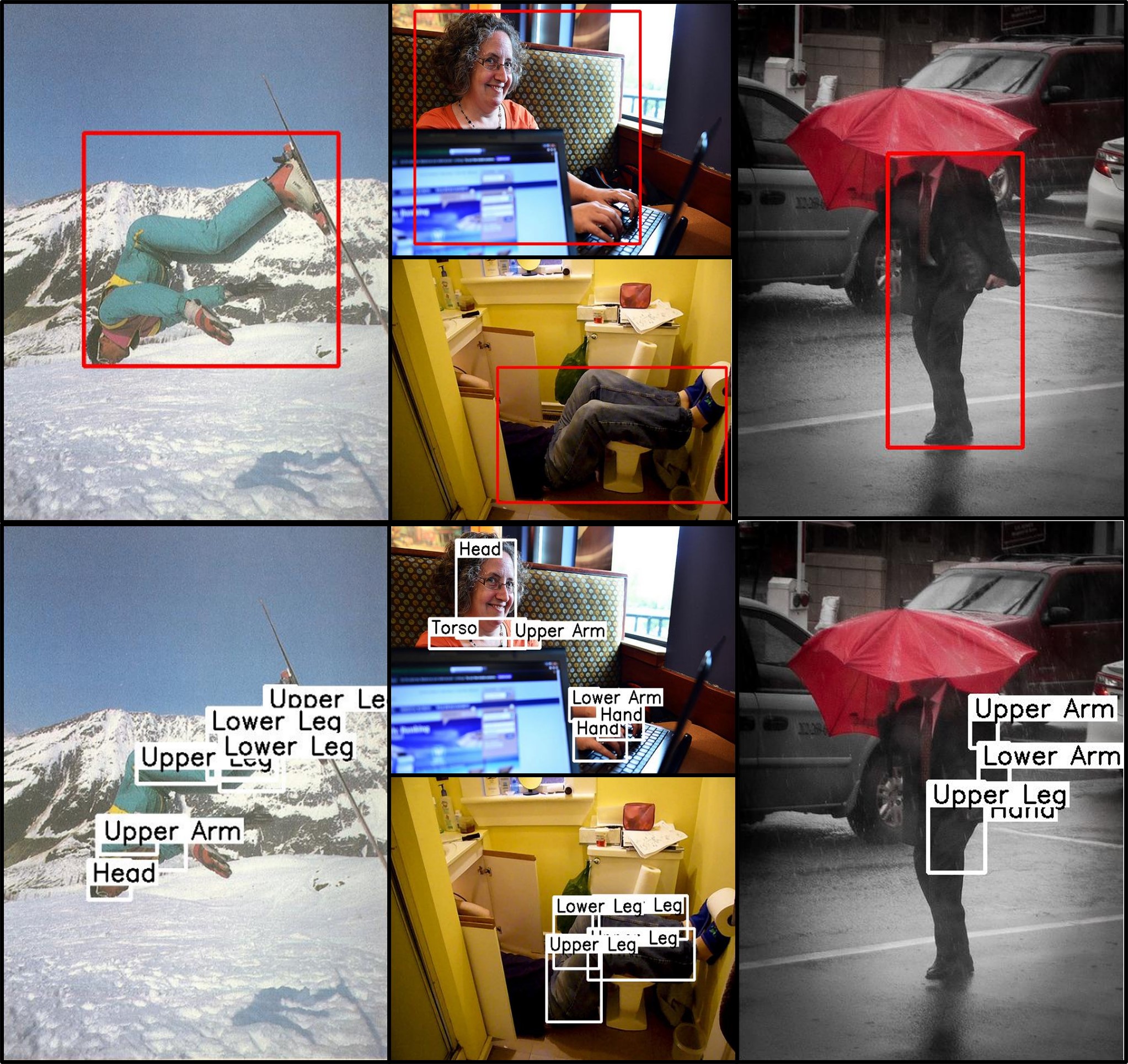}
    \caption{Examples where the person has not been detected by the primary person detector (red boxes), but some body parts could still be detected by the body part detector (white boxes). For the sake of visibility, detection scores have been discarded.}
    \label{fig:coco_fns}
\end{figure}

\begin{figure*}[tb]
      \centering
      \includegraphics[width=\textwidth]{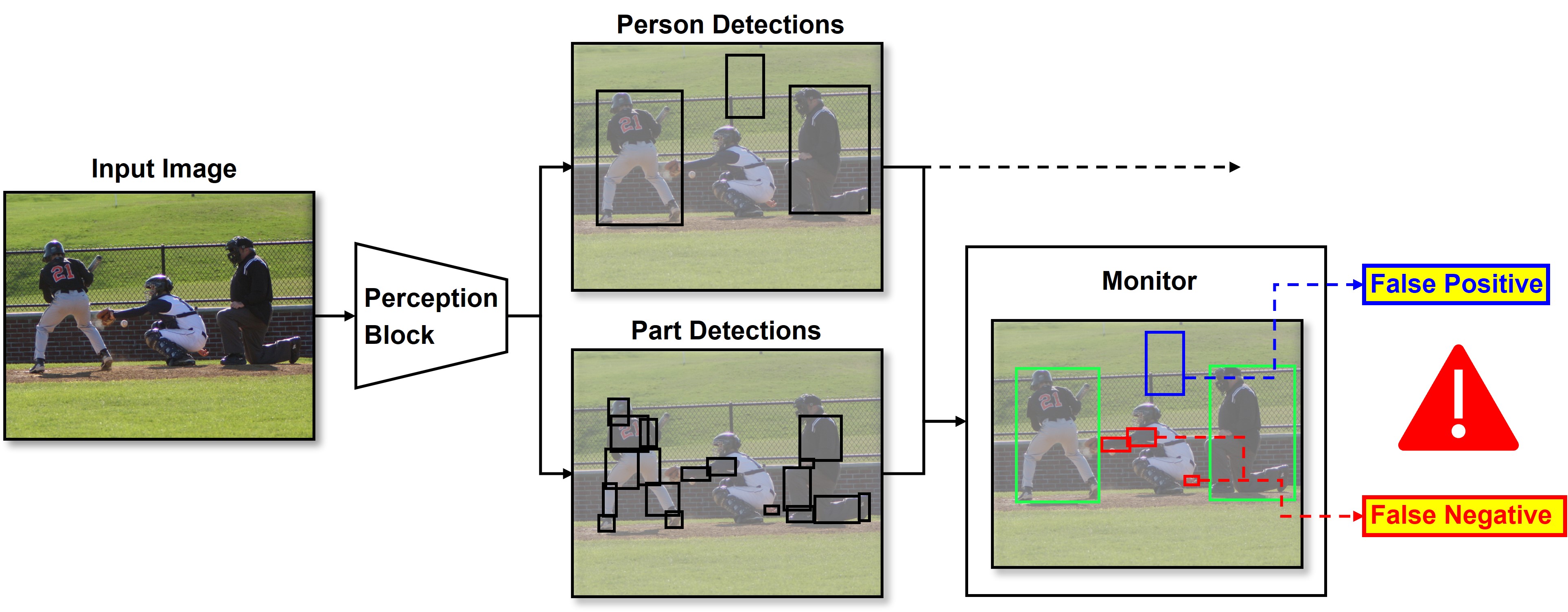}
      \caption{Overview of the \gls{sm}: For an input image, the perception module produces person and body-part detections which are then fed into a monitor performing plausibility checks and raising a warning if inconsistencies between the outputs have been recognized. Green boxes indicate a predicted true positive $d_{person} \in TP_{mon}$, blue boxes a predicted false positive $d_{person} \in FP_{mon}$, and red boxes a predicted false negative $d_{part} \in FN_{mon}$ detection.}
      \label{illustration}
\end{figure*}

2D object detection is a crucial task in computer vision that involves the recognition and localization of various objects of interest in images. Its wide range of applications includes areas such as autonomous systems, medical diagnosis, and agriculture. Although recent advances in deep learning have led to successful object detection models, there remains a challenge in reliably detecting occluded, deformed, or unusually appearing objects \cite{Li2022} at runtime. This is especially important in safety-critical applications, such as autonomous cars or rail vehicles, where preventing incorrect detections is vital. In general, these errors can be roughly divided into two categories: \acrshortpl{fn}, which occur when the model fails to detect present objects, and \acrshortpl{fp}, which occur when the model detects non-existent objects. To address this concern, one approach is to continue improving the performance of the object detector. However, given the limitations of the available training data and the possibility of mistakes during deployment, we argue for the deployment of fault tolerance mechanisms that alert the system to potential errors made by the object detector. These mechanisms could involve re-analyzing the input image or requesting the human to take control of the situation. In this study, we propose a \gls{sm} for object detection that enables the perception module to perform plausibility checks at runtime. Specifically, we focus on person detection, which requires the detection of highly deformable objects with various levels of occlusion and high intra-class variation. The \gls{sm} is based on object detectors that have been trained not only on holistic person objects but also use explicit information about their body parts. The idea behind this approach is that recognizing objects as a whole can sometimes be more difficult than detecting individual object parts, as depicted in Fig.~\ref{fig:coco_fns}. As the human body has a uniform physiological structure (e.g., a head, torso, and limbs), our \gls{sm} takes advantage of this characteristic to cross-check person detections with their detected body parts and alerts the system of potential \acrshort{fp} or \acrshort{fn} errors (see Fig. \ref{illustration}). Although we focus on person detection in this study, our approach could also be applied to other problems, where an object can be characterized by its visual sub-parts.

\textbf{Contributions:} 1) We develop a \gls{sm} for person detection using body parts as additional information source. 2) We propose an evaluation protocol to assess the effectiveness of our proposed monitors in recognizing and localizing errors in the output of the model. 3) We conduct extensive experiments on two benchmark datasets and show that our method yields superior performance in detecting \acrshort{fn} errors compared to our baseline. 4) We find that jointly training on both holistic humans and their individual body parts significantly reduces the amount of initial \acrshort{fp} errors with respect to detecting humans.



\section{RELATED WORK}

\subsection{Runtime Monitoring}

Runtime monitoring is a burgeoning field of research, with various methods falling into three main categories based on their approach to predicting runtime failures \cite{Rahman2021_survey}. The first group of methods relies on previous experience or contextual information to predict failures. For instance, \cite{Zhang2014} developed a system called ALERT that analyzes input data and predicts the likelihood of an unreliable response from a vision system. In \cite{Rahman2019}, a model was trained to detect traffic signs and its internal features were used to train a separate false negative detector. \cite{Rahman2021} focused on the performance difference between training and testing environments and proposed a cascaded neural network to predict the \gls{map} over a sliding window of input frames. Our approach belongs to this category as well, as it utilizes additional more fine-grained information to check for inconsistencies in the person detections. The second group uses methods that recognize inconsistencies by either using temporal or stereo inconsistencies \cite{Ramanagopal2018}, depending on multiple sensor modalities \cite{Zhou2020}, misalignment between the input and output \cite{Kolotouros2019}, or abnormal neural activation pattern \cite{Henzinger2019}, \cite{Cheng2019}. The third group utilizes probabilistic techniques for estimating uncertainty in order to identify low-quality output from the model. Examples include \cite{Feng2020, Harakeh2020}, which estimate uncertainty in object detection models to gauge the reliability of the output. 

\subsection{Part-Based Models}

Early research in object detection \cite{Mohan2011, Mikolajczyk2004} involved training part detectors in a supervised manner and combining their outputs to fit a geometric model. The \gls{dpm} \cite{Felzenszwalb2010} was introduced as a method for handling pose variations by detecting objects as collections of parts localized by local part appearance using HOG \cite{Dalal2005} templates. This was reformulated as a convolutional neural network (CNN) by \cite{Girshick2015}. \cite{Chen2014} addressed the problem of large deformations and partial occlusions in animal detection, modeling objects as collections of body parts and using a loopy graph to capture the spatial and scale relationships between them. \cite{Tian2015} focused on occlusion handling in pedestrian detection and proposed \textit{DeepParts} consisting of an ensemble of CNN-based part classifiers that have been trained on weakly annotated data. In recent research, \cite{Yu2018} proposed a part-based person detector for smartphones and investigated the usefulness of part information. \cite{Gonzalez-Garcia2018} has taken a semantic approach to part detection, using object appearance and class information to guide the detection of parts in the context of their respective objects. \cite{Sitawarin2022} showed that classifying images based on part-segmented objects improves robustness to common corruptions and adversarial attacks.

In light of the availability of datasets with annotations for object parts \cite{Chen2014, r_densepose}, we propose re-examining the role of object part detection in modern object detection systems. Rather than integrating information from the entire object and its parts into a single output, as previously proposed in related publications, we propose to utilize a part detector in addition to the primary task of person detection to monitor the output of the model at runtime. Note that our objective is not to develop a model with superior overall performance, but rather to design a more reliable perception system that comprises modular components.


\section{METHOD}


Let $D_{person}$ denote the set of person detections output by an object detector where each detection has a confidence score above a pre-defined threshold. These pre-defined thresholds are set to achieve a desired operating point for the object detector. For example, one operating point would be to set the confidence thresholds such that we achieve the best trade-off between precision and recall. This can be done by calculating the F1-Score and using the confidence as the threshold where we achieve its maximum. Let $D_{GT}$ denote the set of ground-truth person annotations, then we can classify the predictions made by the object detector into three sets:

\begin{itemize}
\item A set of \acrshortpl{tp} $D_{TP_{gt}}$, including all person detections $d_{person} \in D_{person}$ for which there exists a ground-truth annotation $d_{GT} \in D_{GT}$ such that the \gls{iou} between the bounding boxes is above a pre-defined threshold $\tau$: $\textnormal{IOU}(d_{person}, d_{GT}) > \tau$.

\item A set of \acrshortpl{fp} $D_{FP_{gt}}$, including all person detections $d_{person} \in D_{person}$ for which the previous condition does not hold for any ground-truth annotation $d_{GT} \in D_{GT}$.

\item A set of \acrshortpl{fn} $D_{FN_{gt}}$, including all ground-truth person annotations $d_{GT} \in D_{GT}$ for which it holds $\forall d_{person} \in D_{person}: \textnormal{IOU}(d_{person}, d_{GT}) \le \tau$.
\end{itemize}

In general, one goal in object detection is to minimize the number of \acrshort{fp} and \acrshort{fn} detections while maximizing the number of \acrshort{tp} detections. To accomplish this at runtime, we propose a framework performing plausibility checks to predict the existence and location of \acrshort{fp} and \acrshort{fn} detections. This \gls{sm} is comprised of a monitor that receives both person and body-part detections and decides whether there are inconsistencies between these outputs. Moreover, we also propose an evaluation protocol denoted as \textit{per-object} evaluation to assess how well the monitor can predict and localize false positive or false negatives errors. To the best of our knowledge, this has not been done before. Hence, to show the reasonability of this evaluation protocol, we also designed a reference evaluation protocol denoted as \textit{per-image} evaluation where we re-formulate the problem of error detection as a binary classification task and use established metrics to evaluate the monitors. Therefore, in our studies the monitor operates in two different modes: 

\begin{itemize}
    \item In \textit{per-image} mode, the monitor takes all person and body-part detections $D_{person}$, $D_{part}$ as input and predicts the existence of  at least one \acrshort{fp} or \acrshort{fn} in an image.
    \item In \textit{per-object} mode, the monitor takes all person and body-part detections $D_{person}$, $D_{part}$ as input and predicts a set of \acrshort{tp} $D_{TP_{mon}}$, \acrshort{fp} $D_{FP_{mon}}$, and \acrshort{fn} $D_{FN_{mon}}$ detections.
\end{itemize}

\begin{figure}[tb]
      \centering
      \includegraphics[width=0.45\textwidth]{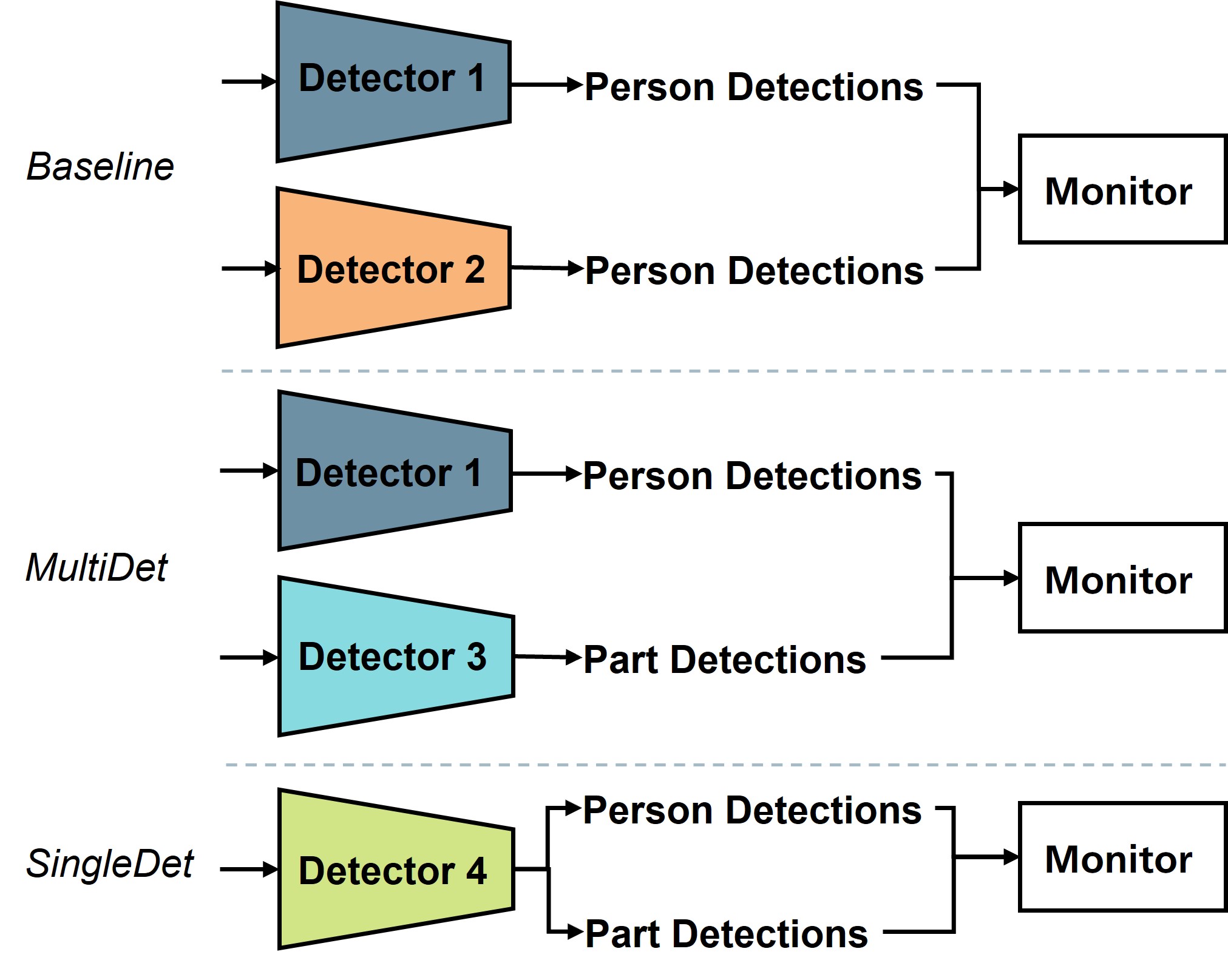}
      \caption{Variants of the \gls{sm}: \textit{Baseline} uses two distinct person detectors that have been trained on class person independently. \textit{MultiDet} also uses two distinct object detectors where one has been trained on class person and the other has been trained on body-part classes. \textit{SingleDet} uses a single object detector that has been trained jointly on class person and body-part classes.}
      \label{methods}
\end{figure}

All considered variants of the \gls{sm} are depicted in Fig.~\ref{methods}: To evaluate the effectiveness of using part detections as additional component for self-monitoring, we compare our approach to a baseline system.  The baseline system referred to as \textit{Baseline} consists of a monitor that receives input from two independently trained person detectors \textit{Detector~1} and \textit{Detector~2}, where \textit{Detector~1} acts as the primary person detection system and \textit{Detector~2} is used to search for inconsistencies in the output. The first variant of our proposed \gls{sm} also consists of two independently trained object detectors \textit{Detector 1} and \textit{Detector 3}, but in contrast to the baseline, the additional object detector \textit{Detector 3} has been trained on the constituent parts of the human body. This system is referred to as \textit{MultiDet} for the rest of this paper. For a fair comparison, both \textit{Baseline} and \textit{MultiDet} use the same primary person detection system \textit{Detector 1}. The second variant of our proposed \gls{sm} consists of a single object detection model \textit{Detector 4} that has been trained jointly on both person and body-part annotations. This system is referred to as \textit{SingleDet} for the rest of this paper.




\section{EXPERIMENTAL SETUP}

\subsection{Datasets}

We trained both person and part detectors on the DensePose \cite{r_densepose} dataset, which is based on the \gls{coco} \cite{r_coco} dataset, and converted the segmentation masks for human body parts into 2D bounding box annotations. We also discarded all images which had body-part annotations for only a subset of humans, resulting in $13,483$ training and $2,215$ validation images. For the evaluation of our proposed \gls{sm}, we used the complete \textit{validation} split of \gls{coco} where humans with a minimum bounding box area of $2,247 \text{ pixels}^2$ are present, which is the minimum area the detectors have seen during training. In total, our \gls{coco} evaluation dataset consists of $11,691$ images. For cross-dataset evaluation, we used the \textit{trainval} split of the \gls{voc} \cite{r_voc} dataset  with a total amount of $2,971$ images after discarding all images where humans with a bounding box area less than $2,247 \text{ pixels}^2$ are present. 

\subsection{Implementation Details}

To evaluate the effectiveness of our approach, we used three different object detection architectures: FCOS \cite{r_fcos} with ResNet-50 \cite{r_resnet} and a \gls{fpn}, YOLOX-S \cite{r_yolox} with Darknet-53, and \gls{rcnn} \cite{r_rcnn} with Resnet-50 as backbone. For training, we used the  default hyperparameter configurations of the open source object detection toolbox \textit{mmdetection} \cite{r_mmdetection} based on PyTorch~\cite{Pytorch:2019}.  All person detectors (\textit{Detector 1} and \textit{2}) for \textit{Baseline} and \textit{MultiDet} have been trained solely on class \textit{person}. The part detectors (\textit{Detector 3}) for \textit{MultiDet} have been trained on a reduced set of DensePose body-part classes where we did not distinguish between left and right, resulting in a total amount of $8$ (Torso, Hand, Foot, Upper Leg, Lower Leg, Upper Arm, Lower Arm, Head) body-part classes. The single model (\textit{Detector 4}) for \textit{SingleDet} has been jointly trained on class \textit{person} and the $8$ body-part classes. The performance for class \textit{person} in terms of \gls{map} and Average Precision (AP) with an \gls{iou} threshold of $0.5$ is presented for each model in Table \ref{tab:single_class_od}. To determine the detection sets derived from the ground-truth annotations $D_{TP_{gt}}$, $D_{FP_{gt}}$, and $D_{FN_{gt}}$, we set the value of the \gls{iou} threshold to $0.5$.

\begin{table}[htb]
\caption{Performance Results in terms of \gls{map} and AP@\gls{iou}=0.5 for the primary person detectors (\textit{Det. 1}, \textit{Det. 4}) on class "person" for Baseline (B), MultiDet (MD), and SingleDet (SD).}
\label{tab:single_class_od}
    \centering
    \begin{tabular}{c c|c c|c c|c c}
    \hline
       \multicolumn{2}{c|}{\multirow{2}{*}{System}} & \multicolumn{2}{c}{\textbf{DensePose}} & \multicolumn{2}{c}{\textbf{COCO}} & \multicolumn{2}{c}{\textbf{PascalVOC}} \\
     & & \footnotesize \textbf{mAP} & $\footnotesize \textbf{AP}_{50}$ & \footnotesize \textbf{mAP} & $\footnotesize \textbf{AP}_{50}$ & \footnotesize \textbf{mAP} & $\footnotesize \textbf{AP}_{50}$ \\
    \hline
    \multirow{2}{*}{fcos} & B/MD & 71.6 & 97.4 & 16.2 & 26.6 & 33.6 & 57.4 \\
     & SD & \textbf{74.0} & \textbf{97.9} & \textbf{19.4} & \textbf{32.6} & \textbf{35.5} & \textbf{60.1} \\
    \hline
    \multirow{2}{*}{yolox} & B/MD & 81.1 & 98.5 & 28.0 & 41.0 & 48.3 & 67.7 \\
    & SD & \textbf{81.4} & \textbf{98.6} & \textbf{30.3} & \textbf{43.0} & \textbf{50.5} & \textbf{69.2} \\
    \hline
    \multirow{2}{*}{c-rcnn} & B/MD & 77.8 & 97.4 & 20.9 & 30.2 & 38.8 & 58.0 \\
    &  SD & \textbf{77.9} & \textbf{98.3} & \textbf{22.3} & \textbf{33.0} & \textbf{41.1} & \textbf{60.8} \\
    \hline
    \end{tabular}
\end{table}


\section{EVALUATION}

\subsection{Per-image Evaluation} \label{sec:image_evaluation}

\subsubsection{Description}

For per-image evaluation, we treated the task of predicting detection errors at runtime as two binary classification problems: For the first classification task, if the monitor predicts the existence of at least one \acrshort{fp} error in the image, it should raise a warning signal (1) and remain silent (0) otherwise. The second classification problem considers \acrshort{fn} errors. The pseudo code for the decision rule is shown in Algorithm \ref{alg:decision_1}. As illustrated in Fig.~\ref{illustration}, a false positive error is considered as a person detection $d_{person} \in D_{person}$ with no associated body-part detection $d_{part} \in D_{part}$. We quantify this person-part association through an overlap threshold $\alpha_{FP}$: For a false positive error, the overlap between the person and all body-part boxes is required to be smaller than $\alpha_{FP} \cdot A_{part}$, where $A_{part}$ is the area of the body-part box. A false negative error is defined as a body part with no associated person detection (c.f. Fig.~\ref{illustration}). Similarly as before, we use an overlap threshold $\alpha_{FN}$ and raise a warning if the person-part overlap is smaller than $\alpha_{FN} \cdot A_{part}$.


Since the monitor returns a binary output without prediction scores, we cannot utilize traditional metrics such as \gls{auroc} and \gls{aupr}. Instead, the evaluation is performed at a specific operating point. This point is determined by two things: Firstly, the confidence thresholds for the object detectors are individually defined by values achieving the maximum F1-Score for the person and body-part classes. Secondly, the values for the overlap thresholds $\alpha_{FP} \in ]0, 1[$ and $\alpha_{FN} \in ]0, 1[$ of the monitor are selected such that they achieve the best performance in terms of \gls{mcc}. We chose \gls{mcc} as the primary metric because it produces a more informative and truthful score in evaluating imbalanced binary classification problems than accuracy~\cite{r_mcc}.

\begin{algorithm}[tb]
	\caption{Decision Rule for Per-Image Evaluation}
	\label{alg:decision_1}
	\begin{algorithmic}[1]
	    \Require{$D_{person}$, $D_{part}$, $\alpha_{FP}$, $\alpha_{FN}$.}
	    \State $ALERT_{FP} \leftarrow \text{False}$, $ALERT_{FN} \leftarrow \text{False}$
		\For {$d_{person}=1,2,\ldots, |D_{person}|$}
			\For {$d_{part}=1,2,\ldots, |D_{part}|$}
			    \State Compute overlap $d_{person} \cap d_{part}$ 
			    \State Compute bounding box area of body part $A_{part}$
			\EndFor
		\EndFor
		\If{$\exists d_{person}: \forall d_{part}: (d_{person} \cap d_{part})  < \alpha_{FP} \cdot A_{part}$}
		    \State $ALERT_{FP} \leftarrow \text{True}$
        \EndIf
		\If{$\exists d_{part}: \forall d_{person}: (d_{person} \cap d_{part})  < \alpha_{FN} \cdot A_{part}$}
		    \State  $ALERT_{FN} \leftarrow \text{True}$
        \EndIf
	\end{algorithmic}
\end{algorithm}


\subsubsection{Results} \label{sec:per_image:results}

Table \ref{tab:experiment:per_image} presents the results for the different monitoring systems \textit{Baseline}, \textit{MultiDet}, and \textit{SingleDet}. For \textit{Baseline}, in Algorithm~\ref{alg:decision_1} $D_{person}$ and $D_{part}$ are determined by the person detections of the underlying two detectors. The results are divided into predicting the existence of at least one false positive $ALERT_{FP} = True$ (left) and one false negative $ALERT_{FN} = True$ (right) error. For each alert type, Table~\ref{tab:experiment:per_image} shows the number of images in which the warning signal was correct or incorrect, i.e. true positive (TP) or false positive (FP), as well as the corresponding precision, recall, and \gls{mcc}. Taking e.g. $ALERT_{FP}$, true positive means that the image indeed contains at least one $FP_{gt}$ detection. As \textit{Baseline} and \textit{MultiDet} share the same primary person detector \textit{Detector~1}, they produce the same number of $FP_{gt}$ and $FN_{gt}$ detections and their performance can be directly compared. In contrast to that, \textit{SingleDet} uses a different person detector that has been jointly trained on person and body-part objects. To highlight this, results for \textit{SingleDet} are presented on a grey background. 


\begin{table*}[tb] 
\caption{Per-Image evaluation results of the three monitoring systems Baseline (B), MultiDet (MD), and SingleDet (SD). Best results are marked in bold.} 

\label{tab:experiment:per_image} 
\centering 
\begin{tabular}{c c|p{6em}|c c c c c||p{6em}|c c c c c}
\hline 
\multicolumn{2}{c|}{\multirow{3}{*}{System}} & Images with $ |FP_{gt}| \ge 1$ & \multicolumn{5}{c||}{$1$: $ALERT_{FP} \leftarrow True$ / $0$: otherwise} & Images with $ |FN_{gt}| \ge 1$ & \multicolumn{5}{c}{$1$: $ALERT_{FN} \leftarrow True$ / $0$: otherwise} \\ 
&  &   & TP & FP & precision & recall & MCC &  & TP & FP  & precision & recall & MCC \\ 
\hline 
 \multicolumn{14}{c}{\textbf{COCO}} \\ 
\hline 
\multirow{3}{*}{fcos} & B & \multirow{2}{*}{1,478 (12.6\%)} &  545 &  751 & 0.42 & 0.37 & 0.31  & \multirow{2}{*}{3,885 (33.2\%)} &  373 &  55 & 0.87 & 0.1 & 0.22 \\  
 &  MD &  &  304 &  117 & 0.72 & 0.21 & \textbf{0.35} &  &  1,698 &  495 & 0.77 & 0.44 & \textbf{0.45} \\  
 &  \cellcolor[gray]{0.8} SD & \cellcolor[gray]{0.8} 943 (8.1\%) & \cellcolor[gray]{0.8} 54 & \cellcolor[gray]{0.8} 49 & \cellcolor[gray]{0.8} 0.52 & \cellcolor[gray]{0.8} 0.06 & \cellcolor[gray]{0.8} 0.15 & \cellcolor[gray]{0.8} 3,834 (32.8\%) & \cellcolor[gray]{0.8} 1,982 & \cellcolor[gray]{0.8} 355 & \cellcolor[gray]{0.8} 0.85 & \cellcolor[gray]{0.8} 0.52 & \cellcolor[gray]{0.8} 0.55 \\  
\hline 
\multirow{3}{*}{yolox} &  B & \multirow{2}{*}{799 (6.8\%)} &  151 &  278 & 0.35 & 0.19 & 0.22 & \multirow{2}{*}{3,221 (27.6\%)} &  646 &  110 & 0.85 & 0.2 & 0.34 \\    
 &  MD &  &  139 &  100 & 0.58 & 0.17 & \textbf{0.29} & &  1,283 &  228 & 0.85 & 0.4 & \textbf{0.49}\\      
  &  \cellcolor[gray]{0.8} SD & \cellcolor[gray]{0.8} 432 (3.7\%) & \cellcolor[gray]{0.8} 9 & \cellcolor[gray]{0.8} 18 & \cellcolor[gray]{0.8} 0.33 & \cellcolor[gray]{0.8} 0.02 & \cellcolor[gray]{0.8} 0.08 & \cellcolor[gray]{0.8} 3,165 (27.1\%) & \cellcolor[gray]{0.8} 1,422 & \cellcolor[gray]{0.8} 274 & \cellcolor[gray]{0.8} 0.84 & \cellcolor[gray]{0.8} 0.45 & \cellcolor[gray]{0.8} 0.53 \\ 
\hline 
\multirow{3}{*}{c-rcnn} &  B & \multirow{2}{*}{1,113 (9.5\%)} &  328 &  455 & 0.42 & 0.29 & \textbf{0.3} & \multirow{2}{*}{3,903 (33.4\%)} &  132 &  20 & 0.87 & 0.03 & 0.13 \\ 
 &  MD &  &  104 &  83 & 0.56 & 0.09 & 0.2 &  &  1,651 &  440 & 0.79 & 0.42 & \textbf{0.45} \\   
 & \cellcolor[gray]{0.8} SD & \cellcolor[gray]{0.8} 639 (5.5\%) & \cellcolor[gray]{0.8}  12 & \cellcolor[gray]{0.8}  28 & \cellcolor[gray]{0.8} 0.3 & \cellcolor[gray]{0.8} 0.02 & \cellcolor[gray]{0.8} 0.06 & \cellcolor[gray]{0.8} 3,975 (34.0\%) & \cellcolor[gray]{0.8} 2,069 & \cellcolor[gray]{0.8} 475 & \cellcolor[gray]{0.8} 0.81 & \cellcolor[gray]{0.8} 0.52 & \cellcolor[gray]{0.8} 0.53 \\   
\hline 
\multicolumn{14}{c}{\textbf{PascalVOC}} \\ 
\hline 
\multirow{3}{*}{fcos} &  B & \multirow{2}{*}{392 (13.2\%)} &  163 &  277 & 0.37 & 0.42 & 0.29 & \multirow{2}{*}{1,039 (35.0\%)}  &  188 &  93 & 0.67 & 0.18 & 0.22 \\ 
 &  MD &  & 74 &  17 & 0.81 & 0.19 & \textbf{0.36} &  &  448 &  158 & 0.74 & 0.43 & \textbf{0.41} \\  
  &  \cellcolor[gray]{0.8} SD & \cellcolor[gray]{0.8} 295 (9.9\%) & \cellcolor[gray]{0.8} 12 & \cellcolor[gray]{0.8} 13 & \cellcolor[gray]{0.8} 0.48 & \cellcolor[gray]{0.8} 0.04 & \cellcolor[gray]{0.8} 0.12 & \cellcolor[gray]{0.8} 1,035 (34.8\%) & \cellcolor[gray]{0.8} 508 & \cellcolor[gray]{0.8} 137 & \cellcolor[gray]{0.8} 0.79 & \cellcolor[gray]{0.8} 0.49 & \cellcolor[gray]{0.8} 0.49 \\ 
\hline 
\multirow{3}{*}{yolox} & B & \multirow{2}{*}{192 (6.5\%)} &  29 &  77 & 0.27 & 0.15 & 0.16 & \multirow{2}{*}{773 (26.0\%)}  & 174 &  46 & 0.79 & 0.23 & 0.34 \\   
 &  MD &  &  28 &  41 & 0.41 & 0.15 & \textbf{0.21} &  &  252 &  71 & 0.78 & 0.33 & \textbf{0.41} \\   
  & \cellcolor[gray]{0.8} SD & \cellcolor[gray]{0.8} 96 (3.2\%) & \cellcolor[gray]{0.8} 3 & \cellcolor[gray]{0.8} 3 & \cellcolor[gray]{0.8} 0.5 & \cellcolor[gray]{0.8} 0.03 & \cellcolor[gray]{0.8} 0.12 & \cellcolor[gray]{0.8} 770 (25.9\%) & \cellcolor[gray]{0.8} 345 & \cellcolor[gray]{0.8} 104 & \cellcolor[gray]{0.8} 0.77 & \cellcolor[gray]{0.8} 0.45 & \cellcolor[gray]{0.8} 0.49 \\  
\hline 
\multirow{3}{*}{c-rcnn} &  B & \multirow{2}{*}{304 (10.2\%)} &  95 &  127 & 0.43 & 0.31 & \textbf{0.31} & \multirow{2}{*}{1,005 (33.8\%)} &  59 &  18 & 0.77 & 0.06 & 0.15 \\ 
 &  MD &  &  22 &  19 & 0.54 & 0.07 & 0.17 & &  377 &  154 & 0.71 & 0.38 & \textbf{0.37} \\     
 &  \cellcolor[gray]{0.8} SD & \cellcolor[gray]{0.8} 174 (5.9\%) & \cellcolor[gray]{0.8} 6 & \cellcolor[gray]{0.8} 13 & \cellcolor[gray]{0.8} 0.32 & \cellcolor[gray]{0.8} 0.03 & \cellcolor[gray]{0.8} 0.09 & \cellcolor[gray]{0.8} 1,046 (35.2\%) & \cellcolor[gray]{0.8} 512 & \cellcolor[gray]{0.8} 170 & \cellcolor[gray]{0.8} 0.75 & \cellcolor[gray]{0.8} 0.49 & \cellcolor[gray]{0.8} 0.46 \\    
\hline 
\end{tabular} 
\vspace{2pt} 
\end{table*}

\textbf{Detecting False Positives}. In terms of \gls{mcc}, both systems \textit{Baseline} ($MCC = 0.16 - 0.31$) and \textit{MultiDet} ($MCC = 0.17 - 0.36$) show a negligible to moderate positive correlation between raising a \acrshort{fp} warning and indeed containing a \acrshort{fp} error in an image on the datasets \gls{coco} and \gls{voc}. These results suggest that the monitors are less suited for predicting \acrshort{fp} errors. In contrast to that, we observe an interesting trend for \textit{SingleDet}: While also having only a negligible relationship ($MCC = 0.06 - 0.15$) between predicting a \acrshort{fp} error and actually containing one, the total number of images with at least one $FP_{gt}$ is significantly reduced by up to $50\%$ in contrast to \textit{Baseline} and \textit{MultiDet}. 





\textbf{Detecting False Negatives.} For all model architectures and on both datasets \gls{coco} and \gls{voc}, \textit{MultiDet} consistently outperforms \textit{Baseline} in terms of \gls{mcc} by a significant margin and exhibits a moderate to strong positive correlation ($MCC= 0.37 - 0.49$) between raising a \acrshort{fn} warning and indeed containing at least one \acrshort{fn} error for that image. In contrast to that, \textit{Baseline} only achieves a negligible to moderate positive correlation ($MCC= 0.13 - 0.34$). \textit{SingleDet} even exhibits a stronger relationship, with \gls{mcc} values reaching from $0.46$ to $0.55$. The results show that using body parts to monitor the output of the person detector helps to predict \acrshort{fn} errors and is superior to only using another person detector to monitor the output. In addition, the results also reveal that we do not need an independent body-part detection component but can achieve even better results with training the same model on both the person class and the body-part classes, which is advantageous to resource-constraint systems.


\subsection{Per-object Evaluation} \label{sec:object_evaluation}

\subsubsection{Description} \label{sec:per_object_description}

In this experiment, we aim to not only predict the existence of errors in images but also pinpoint their locations. To do so, we use a slightly modified version of the monitor decision rule, outlined in Algorithm \ref{alg:decision_2}. Given the sets of person ($D_{person}$) and part detections ($D_{part}$), as well as the overlap thresholds ($\alpha_{FP}, \alpha_{FN}$), the monitor checks for each person detection whether there exists at least one part detection with an intersection of at least $\alpha_{FP} \cdot A_{part}$. If such a part detection is found, the person detection is added to the set of predicted \acrshortpl{tp}  ($D_{TP_{mon}}$); otherwise, it is added to the set of predicted \acrshortpl{fp} ($D_{FP_{mon}}$). A part detection is added to the set of predicted \acrshortpl{fn} ($D_{FN_{mon}}$) if its intersection with all person detections is less than $\alpha_{FN} \cdot A_{part}$. To evaluate the correctness of the predicted sets, we compare them to the sets derived from the ground-truth person annotations ($D_{TP_{gt}}$, $D_{FP_{gt}}$, and $D_{FN_{gt}}$). By doing so, we create a confusion matrix with the following entries:

\begin{itemize}

    \item $(TP_{gt}, TP_{mon})$: \textbf{Correctly detected True Positive.} To determine the number of \acrshortpl{tp} correctly detected by the monitor, we compare the set of \acrshortpl{tp} predicted by the monitor ($D_{TP_{mon}}$) with the ground-truth set ($D_{TP_{gt}}$). If a person detection is present in both sets, it is counted as a correctly detected \acrshort{tp}.
    
    \item $(TP_{gt}, FP_{mon})$: \textbf{Wrongly discarded True Positive.} Likewise, we compare the set of \acrshortpl{fp} predicted by the monitor ($D_{FP_{mon}}$) with the ground-truth set ($D_{TP_{gt}}$). If a ground-truth \acrshort{tp} is present in $D_{FP_{mon}}$, it is counted as a wrongly discarded \acrshort{tp}.
    
    \item $(FP_{gt}, TP_{mon})$: \textbf{Undetected False Positive.} If a ground-truth \acrshort{fp} is not present in $D_{FP_{mon}}$ but present in $D_{TP_{mon}}$ predicted by the monitor, then we count it as undetected.
    
    \item $(FP_{gt}, FP_{mon})$: \textbf{Correctly detected False Positive.} Here, we compare the set of \acrshortpl{fp} provided by the monitor ($D_{FP_{mon}}$) with the ground-truth \acrshortpl{fp} ($D_{FP_{gt}}$). If a \acrshort{fp} is present in both sets, it is counted as a detected \acrshort{fp}.
    
    \item $(FN_{gt}, FN_{mon})$: \textbf{Correctly detected False Negative.} For the detection of \acrshortpl{fn}, a \acrshort{fn} sample $d_{person} \in D_{FN_{gt}}$ is defined to be detected by the monitor if it contains at least one part detection $d_{part} \in D_{FN_{mon}}$:
    $$ \exists d_{part} \in D_{FN_{mon}}: (d_{part} \cap d_{person}) \ge \alpha_{FN} \cdot A_{part} $$
    
    \item $(TN_{gt}, FN_{mon})$: \textbf{Ghost Body Part.} We define ghost body parts as part detections $d_{part} \in D_{FN_{mon}}$ that are not included in any ground-truth annotation $D_{GT}$:
    $$ \forall d_{person} \in {D_{GT}}: (d_{person} \cap d_{part}) < \alpha_{FN} \cdot A_{part} $$
    
\end{itemize}

For evaluation, we consider the following two scenarios: When the focus of our framework lies on detecting \acrshortpl{fp}, then we need to determine whether a person detection might be a ghost detection and discard it. However, this may result in wrongly discarding correct person detections, leading to missed \acrshortpl{tp}. Therefore, we subtract the number of \textbf{wrongly discarded \acrshortpl{tp}} from the number of \textbf{correctly detected \acrshortpl{fp}}. If the balance is positive, then the monitor is useful for \acrshort{fp} detection. If it is negative, this means that the monitor discards more correct person objects than it detects ghost person objects and is therefore harming the overall system. When the focus of our framework lies on detecting \acrshortpl{fn}, then we need to check whether a part detection does not match with any person detection and treat it as part of a \acrshort{fn} detection. However, this also has the effect of potentially producing new ghost detections. Therefore, we subtract the number of \textbf{ghost body parts} from the number of \textbf{correctly detected \acrshortpl{fn}}. Only if the balance is positive, then the monitor is useful for \acrshort{fn} detection.

%
%

\begin{algorithm}[tb]
	\caption{Decision Rule for Per-Object Evaluation}
	\label{alg:decision_2}
	\begin{algorithmic}[1]
	    \Require{$D_{person}$, $D_{part}$, $\alpha_{FP}$, $\alpha_{FN}$.}
	    \State $D_{TP_{mon}} = \varnothing$, $D_{FP_{mon}} = \varnothing$, $D_{FN_{mon}} = \varnothing$
	    
	    \State \textbf{1. Search for fp detections}
		\For {$d_{person}=1,2,\ldots, |D_{person}|$}
		    \If {$\exists d_{part}: (d_{person} \cap d_{part}) \ge \alpha_{FP} \cdot A_{part}$}
		        \State Add $d_{person}$ to $D_{TP_{mon}}$
		    \Else 
		        \State Add $d_{person}$ to $D_{FP_{mon}}$
		    \EndIf
		\EndFor
		
		\State \textbf{2. Search for fn detections}
		\For {$d_{part}=1,2,\ldots, |D_{part}|$}
		    \If {$\forall d_{person}: (d_{person} \cap d_{part}) < \alpha_{FN} \cdot A_{part}$}
		        \State Add $d_{part}$ to $D_{FN_{mon}}$
		    \EndIf
		\EndFor
	\end{algorithmic}
\end{algorithm}

\subsubsection{Results}

Table \ref{tab:experiment:per_object} presents the results of the per-object evaluation in terms of a confusion matrix. Each row of the confusion matrices shows the total number of instances in a detection set $(TP_{gt}, FP_{gt}, FN_{gt}, TN_{gt})$ derived from the ground-truth person annotations, and each column shows the total number of instances in a detection set $(TP_{mon}, FP_{mon}, FN_{mon})$ predicted by the monitor. $(FP_{gt}, FP_{mon})$ and $(FN_{gt}, FN_{mon})$ are shown in green cells, as these combinations stand for a benefit of the monitor. $(TP_{gt}, FP_{mon})$ and $(TN_{gt}, FN_{mon})$ are shown in red cells, as they stand for a decline of the person detection system caused by the monitor. Combinations in white cells are considered to be neutral, as they neither benefit nor harm the overall system by using a monitor. Below each confusion matrix, the previously discussed balances are shown where the values of a red cell are subtracted from the values of a green cell below/above with the focus on \acrshortpl{fp}/\acrshortpl{fn}. 

The \textit{Baseline} method consistently demonstrates a negative balance between correctly detecting \acrshortpl{fp} and wrongly discarding \acrshortpl{tp}. With the exception of YOLOX on \gls{voc}, the balance for \textit{MultiDet} is slightly above $0$, and the results for \textit{SingleDet} exhibit a zero-sum game, displaying a similar trend for predicting \acrshortpl{fp} as observed in Section \ref{sec:per_image:results}. Likewise, a different outcome is observed regarding the balance between detecting \acrshortpl{fn} and producing additional ghost body parts. Compared to \textit{Baseline}, the positive balance for \textit{MultiDet} is up to $9$x higher for \gls{coco} and up to $5$x higher for \gls{voc}, indicating that using a monitor that receives both person and body-part detections effectively helps to identify and localize \acrshort{fn} errors. It should also be noted that the person detectors of \textit{SingleDet} produce significantly fewer \acrshortpl{fp} (up to $50\%$) than the person detectors of \textit{MultiDet} and \textit{Baseline}, consistent with the observations in Section \ref{sec:per_image:results}. Similar to \textit{MultiDet}, a significant positive balance between correctly detecting \acrshortpl{fn} and producing additional ghost detections is observed. This suggests that for resource-constrained systems, it is sufficient to jointly train a single model on person and body parts and use a runtime monitor that searches for inconsistencies within the model output.

\begin{figure}
    \subfigure[\gls{coco}]{
    \centering
    \includegraphics[width=0.47\textwidth]{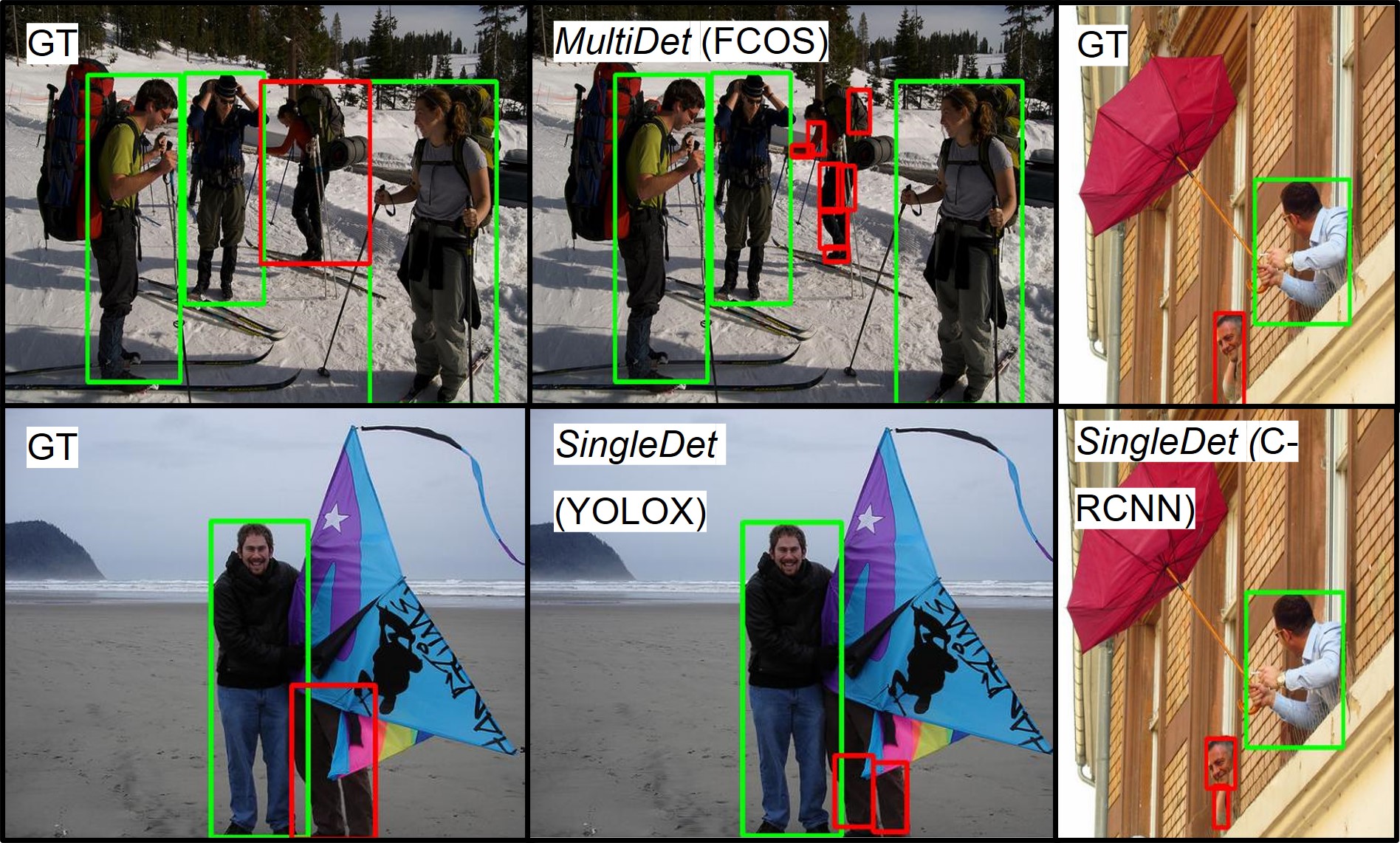}
    } \label{fig:sm_examples_coco}
    \subfigure[\gls{voc}]{
    \centering
    \includegraphics[width=0.47\textwidth]{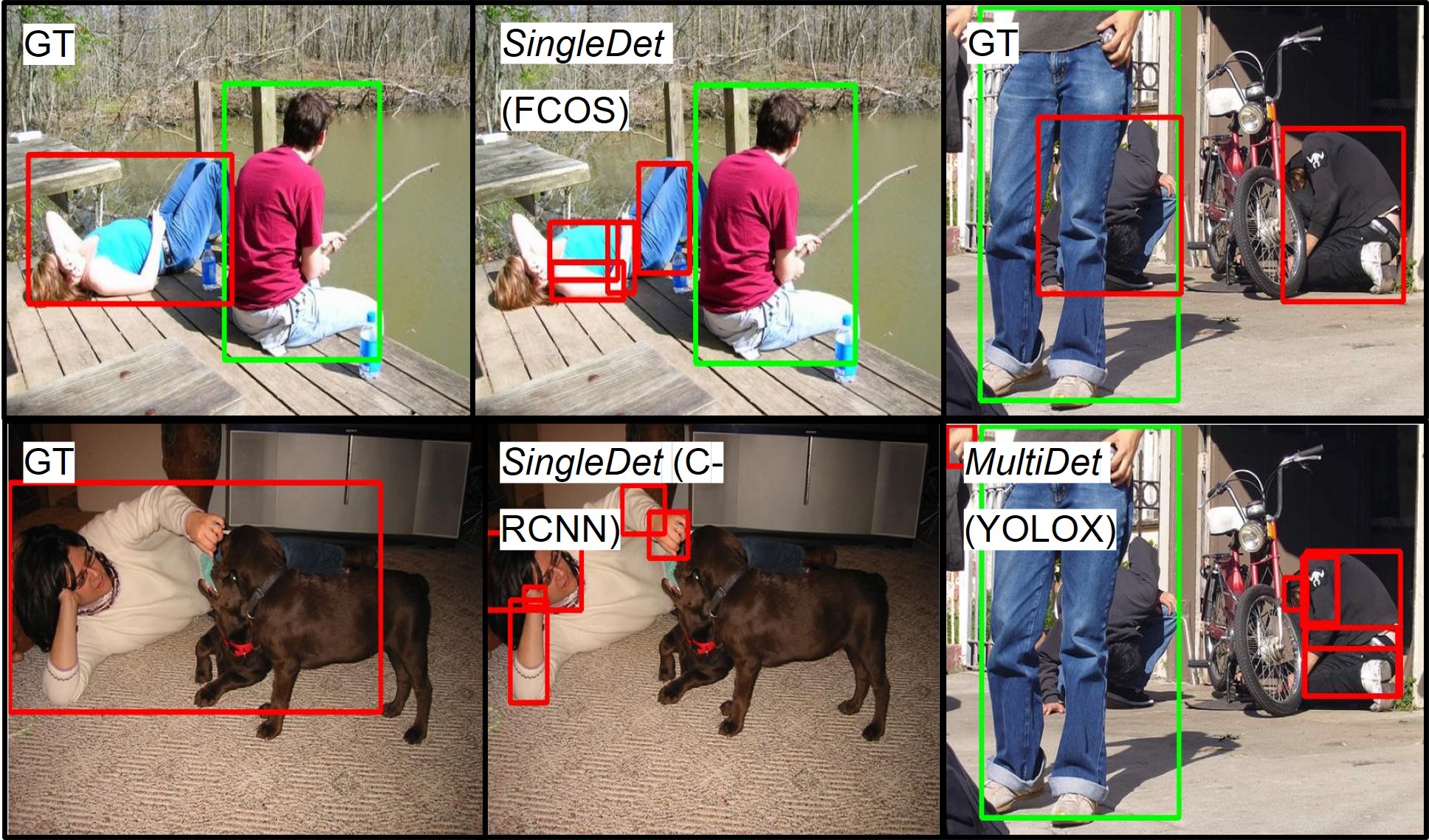}
    } \label{fig:sm_examples_voc}
    \caption{Visualization of the monitoring output. Images marked with \textit{GT} show the detection sets based on ground-truth annotations and images marked with the type of \gls{sm} and model architecture show detection sets produced by the monitor. Green boxes indicate a \acrshort{tp} and red boxes indicate a \acrshort{fn} detection.}
    \label{fig:sm_examples}
\end{figure}

Fig. \ref{fig:sm_examples} shows some qualitative results regarding the monitoring output. Images marked with \textit{GT} show the ground-truth detection sets and images marked with the system name and model architecture (e.g., \textit{SingleDet} (FCOS), \textit{MultiDet} (YOLOX)) show the predicted detection sets produced by the monitor. Green boxes indicate a \acrshort{tp} and red boxes a \acrshort{fn} detection. In the examples, the monitor additionally detected body parts of people that have been missed by the person detector and predicted them as part of a \acrshort{fn} detection.

\begin{table*}[tb]

    \centering
    \caption{Per-object evaluation results: Confusion matrices and balances are shown for the \glspl{sm} Baseline (B), MultiDet (MD), and SingleDet (SD) on the \gls{coco} and \gls{voc} dataset. Best results are marked in bold.}
    \label{tab:experiment:per_object}
    
    \begin{tabular}{c c c}
    
    \hline
    \multicolumn{3}{c}{\textbf{\gls{coco}}} \\
    \hline
    
    \textbf{FCOS} & \textbf{YOLOX} & \textbf{Cascade R-CNN}\\

        \renewcommand\arraystretch{0.7}
        \setlength\tabcolsep{0pt}
    
        \begin{tabular}{c c c c c}
        
            & \MyBoxTop{$TP_{mon}$} & \MyBoxTop{$FP_{mon}$} & \MyBoxTop{$FN_{mon}$} & \\
            
            \MyBox{}{$TP_{gt}$}{}{FFFFFF} & \MyBox{13,326}{13,635}{14,229}{FFFFFF} & \MyBox{447}{138}{46}{FF0000} & \MyBox{}{}{}{000000} & \MyBox{B}{MD}{SD}{FFFFFF}  \\
            
            \MyBox{}{$FP_{gt}$}{}{FFFFFF} & \MyBox{1,260}{1,270}{1,070}{FFFFFF} & \MyBox{319}{309}{48}{00FF00} & \MyBox{}{}{}{000000} & \MyBox{B}{MD}{SD}{FFFFFF} \\
            
            \MyBox{}{$FN_{gt}$}{}{FFFFFF} & \MyBox{}{}{}{000000} & \MyBox{}{}{}{000000} & \MyBox{320}{2,352}{2,487}{00FF00} & \MyBox{B}{MD}{SD}{FFFFFF} \\
            
            \MyBox{}{$TN_{gt}$}{}{FFFFFF} & \MyBox{}{}{}{000000} & \MyBox{}{}{}{000000} & \MyBox{42}{620}{698}{FF0000} & \MyBox{B}{MD}{SD}{FFFFFF} \\
            
        \end{tabular}
        
         &
        \renewcommand\arraystretch{0.7}
        \setlength\tabcolsep{0pt}
        
        \begin{tabular}{c c c c r} 
          
                & \MyBoxTop{$TP_{mon}$} & \MyBoxTop{$FP_{mon}$} & \MyBoxTop{$FN_{mon}$}\\ 
                
               \MyBox{}{$TP_{gt}$}{}{FFFFFF} & \MyBox{15,375}{15,443}{15,562}{FFFFFF} & \MyBox{170}{102}{16}{FF0000} & \MyBox{}{}{}{000000} & \MyBox{B}{MD}{SD}{FFFFFF} \\ 
               
               \MyBox{}{$FP_{gt}$}{}{FFFFFF} & \MyBox{887}{822}{508}{FFFFFF} & \MyBox{81}{146}{8}{00FF00}  & \MyBox{}{}{}{000000} & \MyBox{B}{MD}{SD}{FFFFFF} \\
               
               \MyBox{}{$FN_{gt}$}{}{FFFFFF} & \MyBox{}{}{}{000000} & \MyBox{}{}{}{000000} & \MyBox{574}{1,538}{1,721}{00FF00} & \MyBox{B}{MD}{SD}{FFFFFF} \\
               
               \MyBox{}{$TN_{gt}$}{}{FFFFFF} & \MyBox{}{}{}{000000} & \MyBox{}{}{}{000000} & \MyBox{81}{315}{351}{FF0000} & \MyBox{B}{MD}{SD}{FFFFFF} \\
           
        \end{tabular} &
        
        \renewcommand\arraystretch{0.7}
        \setlength\tabcolsep{0pt}
        
        \begin{tabular}{c c c c r} 
          
                & \MyBoxTop{$TP_{mon}$} & \MyBoxTop{$FP_{mon}$} & \MyBoxTop{$FN_{mon}$} \\ 
                
               \MyBox{}{$TP_{gt}$}{}{FFFFFF} & \MyBox{13,831}{14,084}{13,825}{FFFFFF} & \MyBox{346}{93}{22}{FF0000} & \MyBox{}{}{}{000000} & \MyBox{B}{MD}{SD}{FFFFFF} \\ 
               
               \MyBox{}{$FP_{gt}$}{}{FFFFFF} & \MyBox{1,025}{1,144}{681}{FFFFFF} & \MyBox{224}{105}{10}{00FF00}  & \MyBox{}{}{}{000000} & \MyBox{B}{MD}{SD}{FFFFFF} \\
               
               \MyBox{}{$FN_{gt}$}{}{FFFFFF} & \MyBox{}{}{}{000000} & \MyBox{}{}{}{000000} & \MyBox{138}{1,768}{2,477}{00FF00} & \MyBox{B}{MD}{SD}{FFFFFF} \\
               
               \MyBox{}{$TN_{gt}$}{}{FFFFFF} & \MyBox{}{}{}{000000} & \MyBox{}{}{}{000000} & \MyBox{19}{689}{833}{FF0000} & \MyBox{B}{MD}{SD}{FFFFFF} \\
           
        \end{tabular} \\
        
        & & \\
  
       \begin{tabular}{>{\centering\arraybackslash} m{0.8cm} | >{\centering\arraybackslash} m{1.7cm} | >{\centering\arraybackslash} m{1.7cm}}
           \hline
           \multicolumn{3}{c}{Balances} \\
           \hline
           Method & \textbf{Detecting false positive errors} \tiny $(FP_{gt},FP_{mon}) - (TP_{gt}, FP_{mon})$ & \textbf{Detecting false negative errors} \tiny $(FN_{gt}, FN_{mon}) - (TN_{gt}, FN_{mon})$ \\
           \hline
           B   & -128 & 279 \\
           MD  & \textbf{171} & \textbf{1,732} \\
           \rowcolor[gray]{0.8} SD  & 2 & 1,789 \\
           \hline
    \end{tabular} &
    
    \begin{tabular}{>{\centering\arraybackslash} m{0.8cm} | >{\centering\arraybackslash} m{1.7cm} | >{\centering\arraybackslash} m{1.7cm}}
           \hline
           \multicolumn{3}{c}{Balances} \\
           \hline
           Method & \textbf{Detecting false positive errors} \tiny $(FP_{gt},FP_{mon}) - (TP_{gt}, FP_{mon})$ & \textbf{Detecting false negative errors} \tiny $(FN_{gt}, FN_{mon}) - (TN_{gt}, FN_{mon})$ \\
           \hline
           B   & -89 & 493 \\
           MD  & \textbf{44} & \textbf{1,223} \\
           \rowcolor[gray]{0.8} SD  & -8 & 1,370 \\
           \hline
    \end{tabular} &
    
    \begin{tabular}{>{\centering\arraybackslash} m{0.8cm} | >{\centering\arraybackslash} m{1.7cm} | >{\centering\arraybackslash} m{1.7cm}}
           \hline
           \multicolumn{3}{c}{Balances} \\
           \hline
           Method & \textbf{Detecting false positive errors} \tiny $(FP_{gt},FP_{mon}) - (TP_{gt}, FP_{mon})$ & \textbf{Detecting false negative errors} \tiny $(FN_{gt}, FN_{mon}) - (TN_{gt}, FN_{mon})$ \\
           \hline
           B   & -122 & 119 \\
           MD  & \textbf{12} & \textbf{1,079} \\
           \rowcolor[gray]{0.8} SD  & -12 & 1,644 \\
           \hline
    \end{tabular} \\
    
    & & \\
    \hline
    \multicolumn{3}{c}{\textbf{\gls{voc}}} \\
    \hline
    
    \textbf{FCOS} & \textbf{YOLOX} & \textbf{Cascade R-CNN}\\
        
        \renewcommand\arraystretch{0.7}
        \setlength\tabcolsep{0pt}
        
        \begin{tabular}{c c c c c}
          
                & \MyBoxTop{$TP_{mon}$} & \MyBoxTop{$FP_{mon}$} & \MyBoxTop{$FN_{mon}$} \\ 
                
               \MyBox{}{$TP_{gt}$}{}{FFFFFF} & \MyBox{3,645}{3,761}{3,770}{FFFFFF} & \MyBox{137}{21}{12}{FF0000} & \MyBox{}{}{}{000000} & \MyBox{B}{MD}{SD}{FFFFFF} \\ 
               
               \MyBox{}{$FP_{gt}$}{}{FFFFFF} & \MyBox{355}{354}{318}{FFFFFF} & \MyBox{70}{71}{10}{00FF00}  & \MyBox{}{}{}{000000} & \MyBox{B}{MD}{SD}{FFFFFF} \\
               
               \MyBox{}{$FN_{gt}$}{}{FFFFFF} & \MyBox{}{}{}{000000} & \MyBox{}{}{}{000000} & \MyBox{97}{621}{685}{00FF00} & \MyBox{B}{MD}{SD}{FFFFFF} \\
               
               \MyBox{}{$TN_{gt}$}{}{FFFFFF} & \MyBox{}{}{}{000000} & \MyBox{}{}{}{000000} & \MyBox{17}{250}{288}{FF0000} & \MyBox{B}{MD}{SD}{FFFFFF} \\
           
        \end{tabular} &
    
        \renewcommand\arraystretch{0.7}
        \setlength\tabcolsep{0pt}
        
        \begin{tabular}{c c c c r} 
          
                & \MyBoxTop{$TP_{mon}$} & \MyBoxTop{$FP_{mon}$} & \MyBoxTop{$FN_{mon}$} \\ 
                
               \MyBox{}{$TP_{gt}$}{}{FFFFFF} & \MyBox{4,333}{4,333}{4,340}{FFFFFF} & \MyBox{38}{38}{3}{FF0000} & \MyBox{}{}{}{000000} & \MyBox{B}{MD}{SD}{FFFFFF} \\ 
               
               \MyBox{}{$FP_{gt}$}{}{FFFFFF} & \MyBox{196}{188}{98}{FFFFFF} & \MyBox{14}{22}{3}{00FF00}  & \MyBox{}{}{}{000000} & \MyBox{B}{MD}{SD}{FFFFFF} \\
               
               \MyBox{}{$FN_{gt}$}{}{FFFFFF} & \MyBox{}{}{}{000000} & \MyBox{}{}{}{000000} & \MyBox{139}{370}{413}{00FF00} & \MyBox{B}{MD}{SD}{FFFFFF} \\
               
               \MyBox{}{$TN_{gt}$}{}{FFFFFF} & \MyBox{}{}{}{000000} & \MyBox{}{}{}{000000} & \MyBox{35}{171}{139}{FF0000} & \MyBox{B}{MD}{SD}{FFFFFF} \\
           
        \end{tabular} &
        
        \renewcommand\arraystretch{0.7}
        \setlength\tabcolsep{0pt}
        
        \begin{tabular}{c c c c r} 
          
                & \MyBoxTop{$TP_{mon}$} & \MyBoxTop{$FP_{mon}$} & \MyBoxTop{$FN_{mon}$} \\ 
                
               \MyBox{}{$TP_{gt}$}{}{FFFFFF} & \MyBox{3,850}{3,912}{3,787}{FFFFFF} & \MyBox{82}{20}{13}{FF0000} & \MyBox{}{}{}{000000} & \MyBox{B}{MD}{SD}{FFFFFF} \\ 
               
               \MyBox{}{$FP_{gt}$}{}{FFFFFF} & \MyBox{277}{305}{173}{FFFFFF} & \MyBox{50}{22}{6}{00FF00}  & \MyBox{}{}{}{000000} & \MyBox{B}{MD}{SD}{FFFFFF} \\
               
               \MyBox{}{$FN_{gt}$}{}{FFFFFF} & \MyBox{}{}{}{000000} & \MyBox{}{}{}{000000} & \MyBox{41}{384}{583}{00FF00} & \MyBox{B}{MD}{SD}{FFFFFF} \\
               
               \MyBox{}{$TN_{gt}$}{}{FFFFFF} & \MyBox{}{}{}{000000} & \MyBox{}{}{}{000000} & \MyBox{14}{253}{283}{FF0000} & \MyBox{B}{MD}{SD}{FFFFFF} \\
           
        \end{tabular} \\
        
        & & \\
        
  
      \begin{tabular}{>{\centering\arraybackslash} m{0.8cm} | >{\centering\arraybackslash} m{1.7cm} | >{\centering\arraybackslash} m{1.7cm}}
           \hline
           \multicolumn{3}{c}{Balances} \\
           \hline
           Method & \textbf{Detecting false positive errors} \tiny $(FP_{gt},FP_{mon}) - (TP_{gt}, FP_{mon})$ & \textbf{Detecting false negative errors} \tiny $(FN_{gt}, FN_{mon}) - (TN_{gt}, FN_{mon})$ \\
           \hline
           B   & -67 & 80 \\
           MD  & \textbf{50} & \textbf{371} \\
           \rowcolor[gray]{0.8} SD  & -2 & 397 \\
           \hline
    \end{tabular} &
    
    \begin{tabular}{>{\centering\arraybackslash} m{0.8cm} | >{\centering\arraybackslash} m{1.7cm} | >{\centering\arraybackslash} m{1.7cm}}
           \hline
           \multicolumn{3}{c}{Balances} \\
           \hline
           Method & \textbf{Detecting false positive errors} \tiny $(FP_{gt},FP_{mon}) - (TP_{gt}, FP_{mon})$ & \textbf{Detecting false negative errors} \tiny $(FN_{gt}, FN_{mon}) - (TN_{gt}, FN_{mon})$ \\
           \hline
           B   & -24 & 104 \\
           MD  & \textbf{-16} & \textbf{199} \\
           \rowcolor[gray]{0.8} SD  & 0 & 274 \\
           \hline
    \end{tabular} &
    
    \begin{tabular}{>{\centering\arraybackslash} m{0.8cm} | >{\centering\arraybackslash} m{1.7cm} | >{\centering\arraybackslash} m{1.7cm}}
           \hline
           \multicolumn{3}{c}{Balances} \\
           \hline
           Method & \textbf{Detecting false positive errors} \tiny $(FP_{gt},FP_{mon}) - (TP_{gt}, FP_{mon})$ & \textbf{Detecting false negative errors} \tiny $(FN_{gt}, FN_{mon}) - (TN_{gt}, FN_{mon})$ \\
           \hline
           B   & -32 & 27 \\
           MD  & \textbf{2} & \textbf{131} \\
           \rowcolor[gray]{0.8} SD  & -7 & 300 \\
           \hline
    \end{tabular}
    
    \end{tabular}%

\end{table*}



\section{CONCLUSION AND FUTURE WORK}

In this study, we presented a \gls{sm} for person detection that incorporates a monitor which checks for inconsistencies between the person detection component and an additional body-part detection component in order to alert the system in case of potential errors. Additionally, we developed an evaluation protocol to measure the effectiveness of the monitor in recognizing and localizing errors in the output of the model. Empirical results on \gls{coco} and \gls{voc} have demonstrated that our proposed system using body parts as additional information detects more \acrshort{fn} errors than our baseline system at runtime. In future work, we also plan to explore semi-supervised methods to reduce the cost of labeling object sub-parts. An intriguing outcome of this work is the observation that jointly training an object detection model on the holistic person and the constituent body parts significantly reduces the initial total number of \acrshortpl{fp} with respect to the holistic person class. This raises the hypothesis that training with more fine-grained annotations as an auxiliary task could further improve the model performance, which would be of particular interest in applications where obtaining unlimited data is challenging. One question is also whether this improvement comes from the hierarchy of the human body or whether similar improvements can be also achieved by training on additional classes which are not sub-parts of a human.

\addtolength{\textheight}{-12cm}   

\end{document}